\documentclass[letterpaper]{article} 
\usepackage{aaai23}  
\usepackage{times}  
\usepackage{helvet}  
\usepackage{courier}  
\usepackage[hyphens]{url}  
\usepackage{graphicx} 
\usepackage{natbib}  
\usepackage{caption} 
\usepackage{microtype}

\usepackage{multirow}
\usepackage{makecell}
\usepackage{subfigure}
\usepackage{bm}
\usepackage{booktabs}
\usepackage{url}

\usepackage{amssymb}
\usepackage{amsmath}

\newcommand{\ours}{IFAformer}

\title{On Analyzing the Role of Image for Visual-enhanced Relation Extraction}
\author{
    Lei Li\textsuperscript{\rm 1,2}, 
    Xiang Chen\textsuperscript{\rm 1,2},
    Shuofei Qiao\textsuperscript{\rm 1,2},
    Feiyu Xiong\textsuperscript{\rm 3},
    Huajun Chen\textsuperscript{\rm 1,2},
    Ningyu Zhang\textsuperscript{\rm 1,2}\thanks{Corresponding author}
}
\affiliations{
    \textsuperscript{\rm 1}Zhejiang University \& AZFT Joint Lab for Knowledge Engine, China\\
    \textsuperscript{\rm 2}Hangzhou Innovation Center, Zhejiang University, China\\
    \textsuperscript{\rm 3}Alibaba Group, China\\

    \{leili21,xiang\_chen,shuofei,huajunsir,zhangningyu\}@zju.edu.cn, \\
    feiyu.xfy@alibaba-inc.com
}

\begin{document}

\maketitle

\begin{abstract}
Multimodal relation extraction is an essential task for knowledge graph construction. In this paper, we take an in-depth empirical analysis that indicates the inaccurate information in the visual scene graph leads to poor modal alignment weights, further degrading performance. Moreover, the visual shuffle experiments illustrate that the current approaches may not take full advantage of visual information. Based on the above observation, we further propose a strong baseline with an implicit fine-grained multimodal alignment based on Transformer for multimodal relation extraction. Experimental results demonstrate the better performance of our method\footnote{Codes are available at \url{https://github.com/zjunlp/DeepKE/tree/main/example/re/multimodal}.}.

\end{abstract}

\section{Introduction}

Relation extraction (RE) aims to identify the semantic relations given two entities in a sentence, which plays an essential role in knowledge graph construction and benefits many knowledge-driven tasks.
However, existing mainstream RE methods~\cite{MTB} are text-based and may suffer a sharp performance decline with social media texts since those sentences lack contexts. 
Meanwhile, visual content, such as image posts on Twitter,  often appears alongside the texts. It is intuitive to supplement the missing semantic information with visual content to improve the performance.

Recently, \citet{9428274,multimodal-re} introduce visual-enhanced relation extraction, also known as \textbf{multimodal relation extraction} (MRE), which aims to classify relations between two entities with the assistance of visual contents.
The SOTA method MEGA \cite{multimodal-re} presents an efficient strategy to find the mapping from visual to textual contents based on visual scene graph and syntactic dependency trees, finally improving the MRE performance. 

\begin{figure}[!htbp]
\centering 
\includegraphics[width=0.35\textwidth]{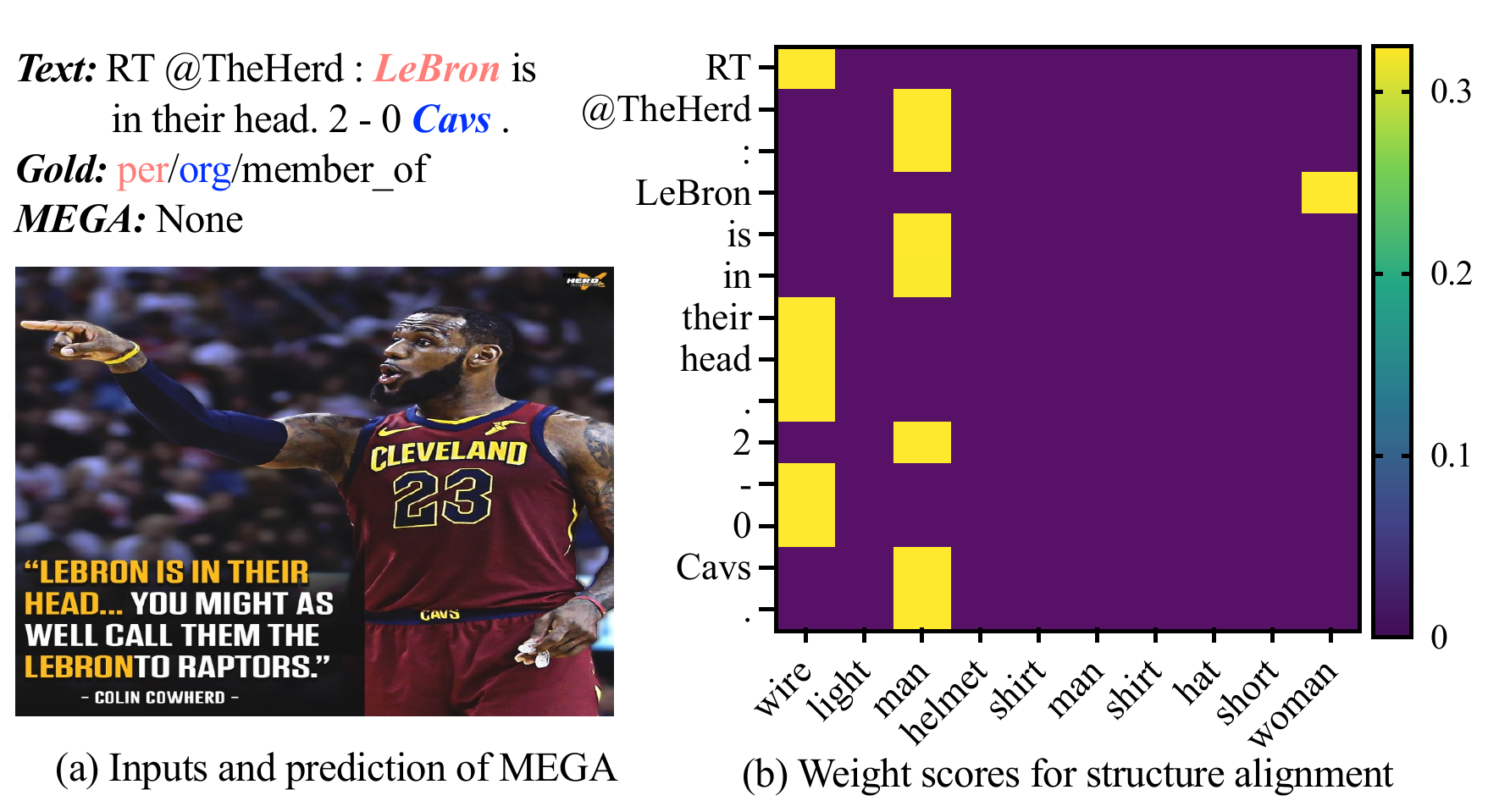} 
\caption{Analysis of the alignment weight scores between visual objects and text words in the MEGA method.} 
\label{fig:mega_weight}
\end{figure}
    
In this paper, we take a deeper look at the internal mechanism of MRE and obtain interesting empirical findings.
We find that not all visual information contributes to performance improvement. 
Moreover, the empirical analysis illustrates that the missing and misleading information in the scene graph may interfere with the decision-making, and the model is not as effective as the pure text-based RE models for some relations.
Building on the above discovery, we work on developing a more realistic visual-enhanced RE model.
We propose an \textbf{I}mplicit \textbf{F}ine-grained multimodal \textbf{A}lignment approach with Trans\textbf{former} ({\ours}), which aligns visual and textual objects in representation space. 



\section{Analysis of the Role of Image for MRE}
\label{sec:vre}

\begin{table*}[htbp!]
\centering
\scalebox{0.75}{
\begin{tabular}{c|cccc|cccc|cccc}
\toprule
\multirow{2}{*}{Method} & \multicolumn{4}{c|}{\textbf{Baseline (standard)}} & \multicolumn{4}{c|}{\textbf{Shuffle (train)}} & \multicolumn{4}{c}{\textbf{Shuffle (test)}} \\ \cline{2-13} 
    & Acc  & Precision  & Recall & F1 
    & Acc  & Precision  & Recall & F1       
    & Acc  & Precision  & Recall & F1  \\ 
\midrule
MTB  & 75.34  & 63.28  & 65.16  & 64.20   
     & 75.34  & 63.28  & 65.16  & 64.20   
     & 75.34  & 63.28  & 65.16  & 64.20   \\
MEGA & 76.15 & 64.51 & 68.44 & 66.41   
     & 76.15  & 65.35  & 64.53  & 64.94   
     & 75.40  & 63.30  & 66.56  & 64.89   \\
     \midrule
{\ours}  & 92.38  & 82.59  & 80.78  & 81.67 
           & 74.23  & 60.84  & 67.97  & 64.21   
           & 47.71  & 29.82  & 29.22  & 29.52 \\
\bottomrule
\end{tabular}
}
\caption{\label{tab:shuffle}
Visual shuffle experiment
on MNRE. Baseline (standard) is the standard unshuffle setting. Shuffle (train) refer to randomly shuffling the images of the training set. {\ours} is the version of Vanilla {\ours} with Visual Objects.
}
\end{table*}


Since in the multimodal relation extraction dataset MNRE ~\cite{multimodal-re}, the improvement of MEGA is relatively small compared with text-based model MTB, we cannot help but question: \emph{Is the ceiling for the improvement of the performance of visual-enhanced RE so low?} 
Thus, we further explore the criticality of visual information in the MRE task. 
We conduct \emph{visual shuffle experiments},  where we randomly shuffle the image-text pairs to compare the performance with the standard dataset.
Then we analyze the structure alignment map between objects and text words of MEGA in Figure \ref{fig:mega_weight}.
We propose the following analysis discussion:

\begin{figure}[!htbp]
\centering 
\includegraphics[width=0.24\textwidth]{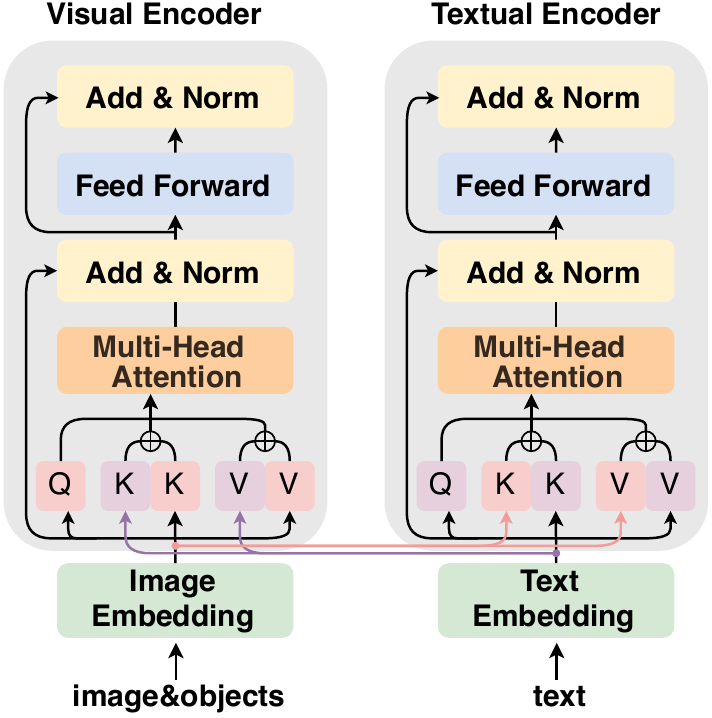} 
\caption{Implicit Fine-grained Multimodal Alignment.} 
\label{fig:model}
\end{figure}

\textbf{Stable or Invalid Performance?}
As shown in Table \ref{tab:shuffle}, although the image-text pairs are mismatched, 
the performance of MEGA does not drop at all.
Besides, we notice that the original MEGA (trained on standard) can obtain stable performance for the mismatched image-text pairs.
Our findings raise doubts about the ability of the model to leverage visual guidance with matched visual features.



\textbf{Is Scene Graph Aligned with Tokens?} 
As shown in Figure \ref{fig:mega_weight}, some visual objects are duplicated (man, shirt) and incorrect (wire), which guides incorrect alignment weights for MEGA. 
We argue these may be attributed to two aspects:
(1) visual objects generated by scene graph are general and straightforward ones, while entities in the text are specific;
(2) coarse-grained alignment cannot fill the semantic gap between visual and text.
Overall, the coarse-grained alignment of scene graph may fail to leverage visual guidance for RE.

\section{A Strong Baseline: IFA}
\label{sec:method}

Based on above observation, we try to mitigate the erroneous graph alignment and propose a strong baseline {\ours}.

\textbf{Multimodal Representation.}
As shown in Figure \ref{fig:model}, we adopt a transformer-based dual-stream architecture to encode multi-modal inputs.
For visual representation, we leverage raw image as the global image and employ visual grounding and object detection to obtain fine-grained subgraphs as local visual objects. 
For textual representation, we tokenize the text into token sequence, then feed it into the textual encoder.

\textbf{Implicit Fine-grained Multimodal Alignment.}
We apply an implicit token-object alignment via multi-granularity signals at each layer of encoders to capture correlation between visual objects and entities. 
Given the hidden features $\bm{H}^l_t, \bm{H}^l_v \in {\mathbb{R}^{n\times d}}$ at the $l$-th layer of text and visual encoder, respectively.
We project them into query/key/value vector:
\begin{equation}
\small
\label{eq:qkv}
    \bm{Q}^l, \bm{K}^l, \bm{V}^l = \bm{x}\bm{W}^l_q, \bm{x}\bm{W}^l_k, \bm{x}\bm{W}^l_v; \bm{x} \in \{\bm{H}^l_t, \bm{H}^l_v\}
\end{equation}
where $\bm{W}^l_q, \bm{W}^l_k, \bm{W}^l_v \in{\mathbb{R}^{d\times d_h}} $ are attention projection parameters. 
Then the hidden features at ($l$+1)-th layer through the multi-head attention can be calculated as follows:
\begin{equation}
\small
\begin{aligned}
    \label{eq:attn}
    \bm{H}^{l+1}_t = \mathrm{Attn}(\bm{Q}^l_t, [\bm{K}^l_v, \bm{K}^l_t], [\bm{V}^l_v, \bm{V}^l_t])\\
    \bm{H}^{l+1}_v = \mathrm{Attn}(\bm{Q}^l_v, [\bm{K}^l_t, \bm{K}^l_v], [\bm{V}^l_t, \bm{V}^l_v])
\end{aligned}
\end{equation}
Lastly, we use the output of textual encoder to do prediction.

\section{Experiments}

\begin{table}[t!]
\centering
\small
\scalebox{0.8}{
\begin{tabular}{l|cccc}
\toprule

Methods
& Acc & Precision  & Recall  & F1 \\


\midrule
    PCNN* & 73.36 & 69.14 & 43.75 & 53.59
    \\
    BERT* & 71.13  & 58.51 & 60.16 & 59.32
    \\
    MTB* & 75.34 & 63.28 & 65.16 & 64.20
    \\
\midrule

 
    BERT+SG+Att. & 74.59  & 60.97 & 66.56 & 63.64
    \\
    ViLBERT & 74.89 & 64.50 & 61.86 & 63.61\\
    MEGA & 76.15 & 64.51 & 68.44 & 66.41  
    \\
\cmidrule{1-5}

    Vanilla \ours & 87.75 & 69.90 & 68.11 & 68.99 \\
    \enspace w/o Text Attn. & 76.21 & 66.95 & 61.72 & 64.23 \\
    

    \enspace w/ Visual Objects & \textbf{92.38}
     & \textbf{82.59} & \textbf{80.78} & \textbf{81.67} \\

\bottomrule
\end{tabular}
}
\caption{\label{tab:result1}
The overall performance of baselines on MNRE.  
}
\end{table}

The overall results can be seen in Table \ref{tab:result1}.
We observe that our Vanilla {\ours} is superior to all text-based models and the newest SOTA model MEGA.
From Table \ref{tab:shuffle}, we can find the performance of  {\ours}  drops significantly when disrupting the image-text pairs, revealing that our model indeed utilizes the visual information for multimodal relation extraction.
In addition, we conduct an ablation study and observe that: 
(1) \textit{w/o Text Attn.}: Removing textual information in the visual encoder will reduce performance, revealing that fusing multiple information plays a vital role in MRE.
(2) \textit{w/ Visual Objects.}: Incorporating fine-grained visual objects as described in the Method section significantly improve performance and outperforms MEGA 15.26\% F1 scores, indicating the significance of the fine-grained visual features for MRE.



    

\section{Conclusion}
In this paper, we study MRE and take an in-depth empirical analysis that indicates the pain points of current MRE methods. 
We further propose a strong baseline {\ours}.
Experimental results demonstrate the effectiveness.

\section*{Acknowledgment}

This work was supported by the National Natural Science Foundation of China (No.62206246), Zhejiang Provincial Natural Science Foundation of China (No. LGG22F030011), Ningbo Natural Science Foundation (2021J190), Yongjiang Talent Introduction Programme (2021A-156-G), and CAAI-Huawei MindSpore Open Fund.

\bibliography{aaai23}

\end{document}